\documentclass[letterpaper, 10 pt, conference]{ieeeconf}  

\IEEEoverridecommandlockouts                              

\usepackage[utf8]{inputenc}
\usepackage[T1]{fontenc}
\usepackage{amsfonts}
\usepackage{amsmath}
\usepackage{booktabs}
\usepackage{caption}
\usepackage{color}
\usepackage{courier}
\usepackage{graphicx}
\usepackage{adjustbox}
\usepackage{makecell}
\usepackage{microtype}
\usepackage{multirow,makecell}
\usepackage{nicefrac}
\usepackage{subcaption}
\usepackage{url}
\usepackage{xcolor}
\usepackage{hyperref}
\usepackage{xcolor,colortbl}
\usepackage{pgfplots}
\newif\ifdark
\IfFileExists{/Users/vedaldi/.bash_profile}{
\makeatletter
\immediate\write18{\unexpanded{%
if defaults read -g AppleInterfaceStyle 2>/dev/null;
then echo \\darktrue > /tmp/displaymode.tex;
else echo \\darkfalse > /tmp/displaymode.tex; fi}}
\makeatother 
\input{/tmp/displaymode.tex}
\ifdark
\definecolor{pcolor}{HTML}{1E1E1E}
\definecolor{tcolor}{HTML}{C5C5C5}
\else
\definecolor{pcolor}{HTML}{FDF6E3}
\definecolor{tcolor}{HTML}{333333}
\fi
\pagecolor{pcolor}
\color{tcolor}
\hbadness=\maxdimen
\vbadness=\maxdimen
\vfuzz=30pt
\hfuzz=30pt
}{} 
\pgfplotsset{compat=1.17}
\usepackage{float}
\usepackage{stfloats}
\usepackage{cleveref}
\makeatletter
\renewcommand{\paragraph}{%
  \@startsection{paragraph}{4}%
  {\z@}{0em}{-1em}%
  {\normalfont\normalsize\bfseries}%
}
\makeatother

\crefname{figure}{Fig.}{Fig.}
\Crefname{figure}{Fig.}{Fig.}
\crefformat{equation}{(#2#1#3)}

\title{\LARGE \bf
Lifting 2D Object Locations to 3D by Discounting LiDAR Outliers across Objects and Views
}

\author{Robert McCraith$^{1}$, Eldar Insafutdinov$^{2}$, Lukas Neumann$^{3}$, Andrea Vedaldi$^{4}$

\thanks{
    $^{1,2,3,4}$ Authors are with Visual Geometry Group, Dept.~of Engineering Science, University of Oxford
    {\tt\small \{robert,eldar, lukas,vedaldi\}@robots.ox.ac.uk}
}
}

\begin{document}

\maketitle
\thispagestyle{empty}
\pagestyle{empty}

\begin{abstract}
We present a system for automatic converting of 2D mask object predictions and raw LiDAR point clouds into full 3D bounding boxes of objects.
Because the LiDAR point clouds are partial, directly fitting bounding boxes to the point clouds is meaningless.
Instead, we suggest that obtaining good results requires sharing information between \emph{all} objects in the dataset jointly, over multiple frames.
We then make three improvements to the baseline.
First, we address ambiguities in predicting the object rotations via direct optimization in this space while still backpropagating rotation prediction through the model.
Second, we explicitly model outliers and task the network with learning their typical patterns, thus better discounting them.
Third, we enforce temporal consistency when video data is available.
With these contributions, our method significantly outperforms previous work despite the fact that those methods use significantly more complex pipelines, 3D models and additional human-annotated external sources of prior information.
\end{abstract}

\section{Introduction}\label{s:introduction}

Robotics applications often require to recover the 3D shape and location of objects in world coordinates.
This explains the proliferation of datasets such as KITTI, nuScenes and SUN RGB-D~\cite{geiger12are-we-ready,nuscenes2019,sun2020scalability} which allow to train models that can classify, detect and reconstruct objects in 3D from sensors such as cameras and LiDARs.
However, creating such datasets is very expensive.
For example,~\cite{sun-rgbd,wang2019apolloscape} report that manually annotating a single object with a 3D bounding box requires approximately 100 seconds.
While this cost has since been reduced~\cite{meng2020ws3d}, it remains a significant bottleneck in data collection.

\begin{figure}[h]
    \centering
    \begin{tabular}{c c}
        \includegraphics[width=0.5\textwidth]{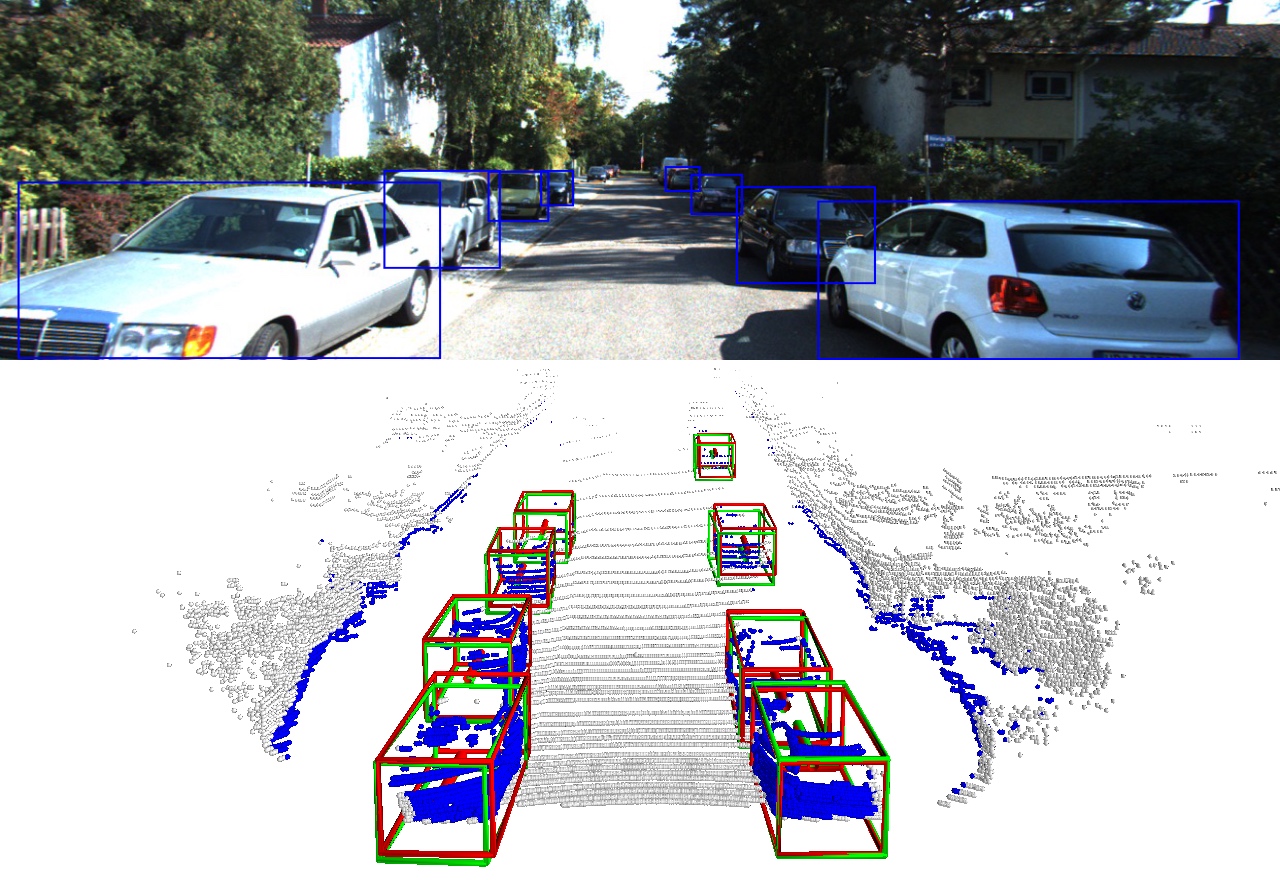} 
        \\ & 
         
        
 
    \vspace{-15pt}

    \end{tabular}
    \caption{Our Method (green) vs KITTI label (red), no Ground Truth is used to train our model.\vspace{-10pt} }
    \label{f:Promo}

\end{figure}
In some cases, cross-modal learning can substitute manual annotations.
An example is monocular depth prediction, where supervision from a LiDAR sensor is generally sufficient~\cite{Fu_2018_CVPR}.
Unfortunately, this does not extend to tasks such as object detection.
For instance, a dataset such as KITTI provides only $7481$ video frames annotated with objects due to the cost of manual annotation.
In this work, we thus consider the problem of detecting objects in 3D, thus also automatically annotating them in 3D, but using only standard 2D object detector trained on a generic dataset such as MS-COCO~\cite{lin2014microsoft}, which is disjoint from the task-specific dataset (in our case KITTI for 3D car detection).
We assume to have as input a collection of video frames, the corresponding LiDAR readings from the viewpoint of a moving vehicle, and the ego-motion of the vehicle.
We also assume to have a pre-trained 2D detector and segmenter such as Mask R-CNN\@ for the objects of interest (e.g., cars).
With this information, we wish to train a model which takes in raw LiDAR point cloud and outputs full 3D bounding box annotation for the detected objects without incurring any further manual annotation cost.


There are three main challenges.
First, due to self and mutual occlusions, the LiDAR point clouds only cover part of the objects in a context dependent manner.
Second, the LiDAR readings are noisy, for example sometimes seeing through glass surfaces (windows) and sometimes not.
Third, the available 2D segmentations may not be perfectly semantically aligned with the target class (e.g., `vehicles' vs `cars'), are affected by a domain shift, and may not be perfectly geometrically with the LiDAR data, resulting in a large number of outlier 3D points arising from background objects.

We propose an approach based on the following key ideas.
First, because the LiDAR point cloud can only cover the object partially, it is impossible to estimate the full 3D extent of the object from a single observation of it.
Instead, we share information between all predicted 3D boxes in the dataset by \emph{learning a 3D bounding box predictor from all the available data}.
We further aid the process by injecting weak prior information in the form of a single fixed 3D mesh template of the object (an `average car'), but avoid sophisticated 3D priors employed in prior works~\cite{sdflabel,qin20weakly}.

We then introduce three improvements to the `obvious' baseline implementation of this idea.
First, we show that a key challenge in obtaining good 3D bounding boxes is to estimate correctly the yaw (rotation) of the object.
This is particularly challenging for partial point clouds as several ambiguous fits (generally up to rotations of 90 degrees) often exist.
Prior work has addressed this problem by using pre-trained yaw predictors, requiring manual annotation.
Here, we learn the predictor automatically from the available data only.
To this end, we show that mere local optimization via gradient descent works poorly;
instead, we propose to systematically explore a full range of possible rotations for each prediction, backpropagating the best choice every time.
We show that this selection process is very effective at escaping local optima and results in excellent, and automated, yaw prediction.

Second, we note that the LiDAR data contains significant outliers.
We thus propose to automatically learn the \emph{pattern} of such outliers by predicting a confidence score for each 3D point, treated as a Gaussian observation.
These confidences are self-calibrated using a neural network similar to the ones used for point cloud segmentation, configured to model the aleatoric uncertainty of the predictor.

Third, we note that we generally have at our disposal \emph{video data}, which contains significant more information than instantaneous observations.
We leverage this information by enforcing a simple form of temporal consistency across several frames.

As a result of these contributions we obtain a very effective system for automatically labelling 3D objects.
Our system is shown empirically to outperform relevant prior work by a large margin, all the while being simpler, because it uses less sources of supervision and because it does \emph{not} use sophisticated prior models of the 3D objects nor a large number of 3D models as priors~\cite{sdflabel,qin20weakly}

\section{Related work}\label{s:related}



\paragraph{Supervised}

3D object detection methods assume availability of either monocular RGB images, LiDAR point clouds or both. Here we focus on supervised methods using 3D point cloud inputs. \cite{wang2015voting,engelcke2017vote3deep} discretise point clouds onto a sparse 3D feature grid and apply convolutions while excluding empty cells. \cite{chen2017multiview} project the point cloud onto the frontal and the top views, apply 2D convolutions thereafter and generate 3D proposals with an RPN \cite{ren16faster}. \cite{zhou2018voxelnet} convert the input point cloud into a voxel feature grid, apply a PointNet \cite{qi16pointnet:} to each cell and subsequently process it with a 3D fully convolutional network with an RPN head which generates object detections.
Frustum PointNets \cite{qi2017frustum} is a two step detector which first detects 2D bounding boxes using these to determine LiDAR points of interest which are filtered further by a segmentation network. The remaining points are then used to infer the 3D box parameters with the centre prediction being simplified by some intermediate transformations in point cloud origins. We are using Frustum PointNets as a backbone for our method.

\paragraph{Weakly Supervised}

Owing to the complexity of acquiring a large scale annotated dataset for 3D object detection many works recently have attempted to solve this problem with less supervision. \cite{meng2020ws3d} the required supervision is reduced from the typical $(X,Y,Z)$ centre, $yaw$, $(x_1,y_1,x_2,y_2)$ 2D box and $(l,w,h)$ 3D size they instead annotate 500 frames with centre $(X,Z)$ in the Birds Eye View and finely annotating a 534 car subset of these frames to achieve accuracy similar to models trained on the entire Kitti training set. This however can result in examples in the larger training set or validation set which are outside the distribution of cars seen in the smaller subset, are also susceptible to the problems mentioned in \cite{feng2020labels} and the weakly annotated centres can have a large difference to the Kitti annotations. In \cite{sdflabel} trains DeepSDF\cite{Park_2019_CVPR} on models from a synthetic dataset which are then rendered into the image and iteratively refined to produce a prediction that best fits the predictions of Mask R-CNN outputs.
This however takes 6 seconds to refine predictions on each input example and ground truth 2D boxes are used to select cars used to train on. In \cite{qin20weakly} 3D anchors densely placed across the range of annotations are projected into the image with object proposed by looking at the density of points within or nearby the anchors in 3D and the 3D pose and 2D detections are supervised by a CNN trained on Beyond PASCAL\cite{xiang14beyond}. In \cite{koestler2020learning} instance segmentation is used to place a mesh in the location of detected cars which is refined using supervised depth estimation. 

\paragraph{Viewpoint estimation}
Estimating 3D camera orientation is an actively researched topic \cite{pepik12teaching, kendall2015posenet, tulsiani15viewpoints, su2015render, mousavian20173d, prokudin2018deep, liao2019spherical}. Central to this problem is the choice of an appropriate representation for rotations. Directly representing rotations as angles $\theta \in [0, 2\pi )$ suffers from discontinuity at $2\pi$. One way to mitigate this is to use the trigonometric representation $(\cos \theta, \sin \theta)$ \cite{penedones2012improving, massa2016crafting, beyer2015biternion}. Another approach is to discretise the angle into $n$ quantised bins and treat viewpoint prediction as a classification problem \cite{tulsiani15viewpoints, su2015render, mousavian20173d}. Quaternions are another popular representation of rotations used with neural networks \cite{kendall2015posenet, xiang2017posecnn}. Learning camera pose without direct supervision, for example, when fitting a 3D model to 2D observations, may suffer from ambiguities due to the object symmetries \cite{saxena2009learning}. Practically, this means that the network can get stuck in a local optima of $SO(3)$ space and not be able to recover the correct orientation. Several recent works \cite{tulsiani2018multi, insafutdinov18pointclouds, goel20shape} overcome this by maintaining several diverse candidates for the estimated camera rotation and selecting the one that yields the best reconstruction loss. Our approach discussed in sec.~\ref{s:direct-yaw} is most related to these methods.

\newcommand{\X}{\mathbf{X}}
\section{Method}\label{s:method}

\begin{figure*}
\centering
\includegraphics[width=1.0\textwidth]{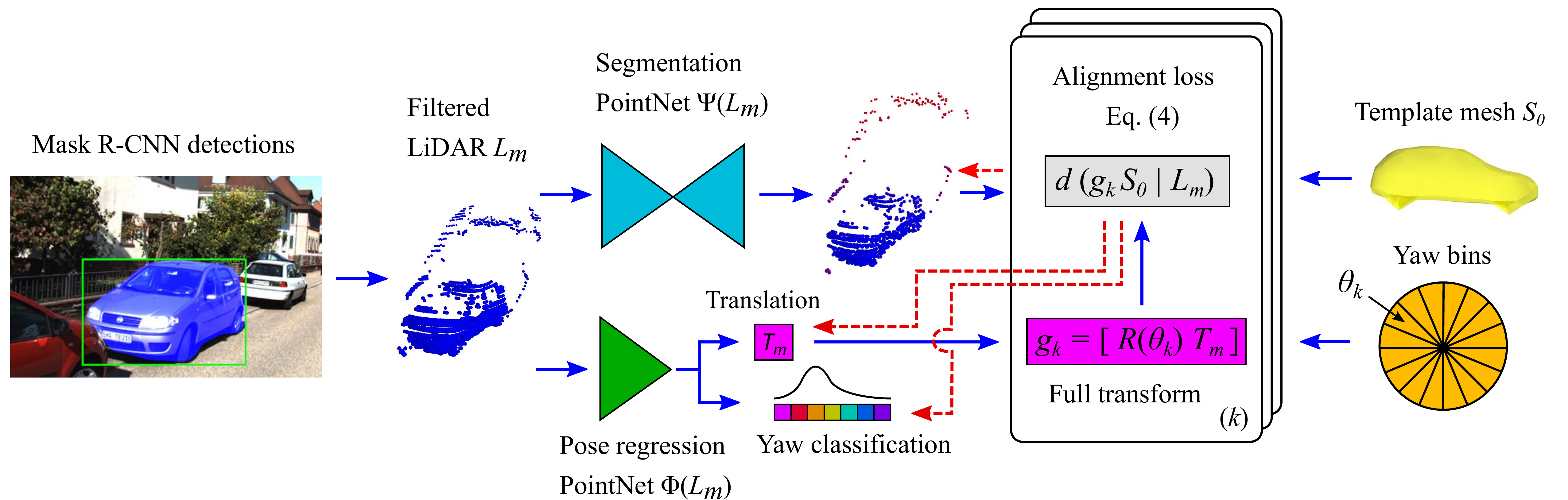}
\caption{Network architecture. Dashed arrows in red show the flow of the gradients during the backward pass. Alignment loss is evaluated for each yaw bin $\theta_k$ and the optimal yaw is used to supervise the yaw classification network $\Phi_r$ (see sec.~\ref{s:direct-yaw}).\vspace{-15pt}}\label{fig:network_architecture}

\end{figure*}

Our goal is to estimate the 3D bounding box of objects given as input videos with 2D mask predictions and LiDAR point clouds.
We discuss first how this problem can be approached by direct fitting and then develop a much better learning-based solution.

\subsection{Shape model fitting}\label{s:shapemodel}

We first describe how a 3D bounding box can be fitted to the available data in a direct manner.
To this end, let $I \in \mathbb{R}^{3\times H\times W}$ be a RGB image obtained from the camera sensor and let $L \subset\mathbb{R}^3$ be a corresponding finite collection of 3D points $\X\in L$ extracted from the LiDAR sensor.
Furthermore, let $m \in \{0,1\}^{H \times W}$ be the 2D mask of the object obtained from a system such as Mask R-CNN~\cite{he17mask} from image $I$.
Our goal is to convert the 2D mask $m$ into a corresponding 3D bounding box $B$.

\begin{figure}[b]
    \centering
    \begin{tabular}{c c c c}
         \includegraphics[width=0.25\textwidth]{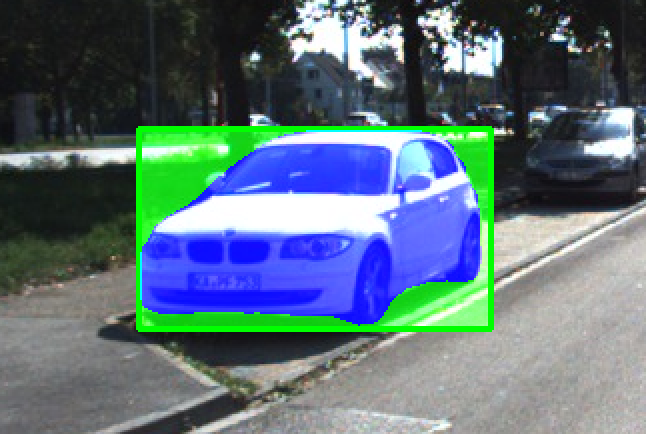} &
         \includegraphics[width=0.2\textwidth]{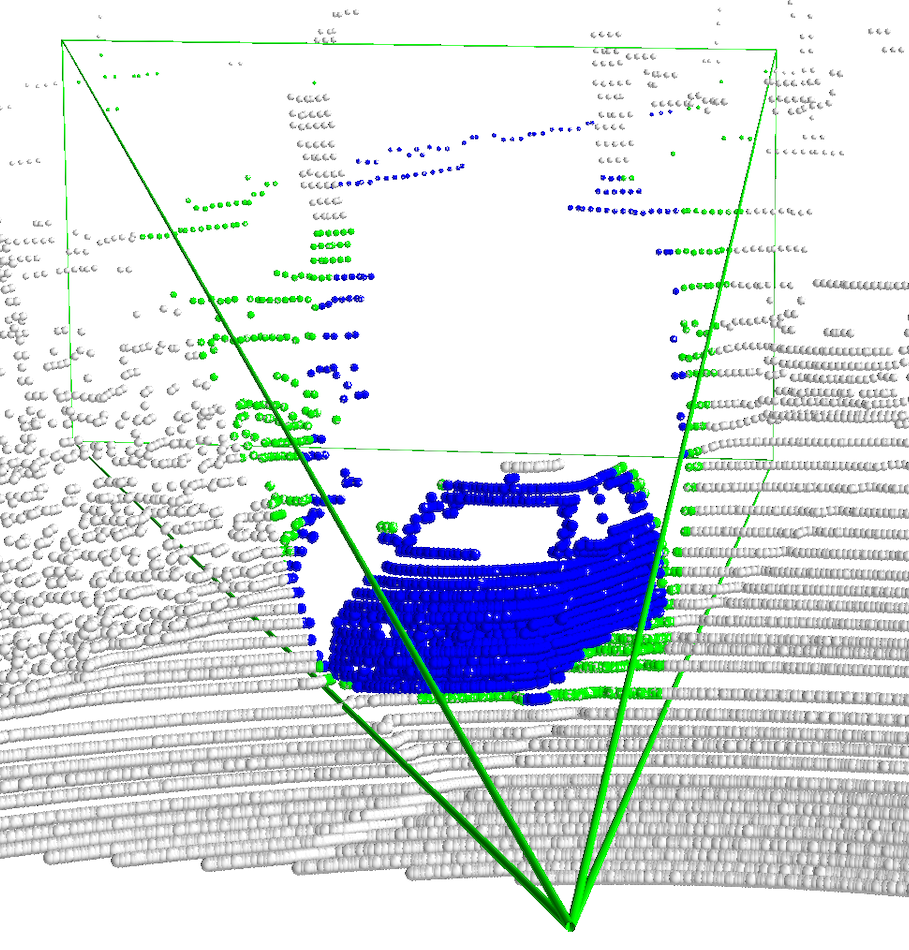} 
    \end{tabular}
    \caption{
    For each pair, left: RGB input image with Mask R-CNN predicted box and highlighted pixels inside the mask.
    Right: LiDAR points in blue represent those inside the 2D mask, green those outside.
    Note that, while the image mask removes many outliers, many remain at the object boundaries and transparent surfaces.}\label{fig:dataset_example}
\end{figure}

To do so, assume that the LiDAR points are expressed in the reference frame of the camera and that the camera calibration function $k : \mathbb{R}^2 \rightarrow \Omega = \{1,\dots,H\}\times\{1,\dots,W\}$ is known.
The calibration function is defined such that the 3D point $\X=(X,Y,Z)$ projects onto the image pixel $u = k(\pi(\X))$ where $\pi(X,Y,Z)=(X/Z,Y/Z)$ is the perspective projection.
In particular, the subset of LiDAR points $L_m \subset L$ that project onto the 2D mask $m$ is given by:
$
L_m
=
L \cap
(k \circ \pi)^*(m)
$
where ${}^*$ denotes the pre-image of a function.
In practice, this is a crude filtering step, because the masks are imprecise and not perfectly aligned to the LiDAR and because LiDAR may sometimes see `through' the object, for instance in correspondence of glass surfaces
(see \Cref{fig:dataset_example}).

In order to fit a 3D bounding box $B$ to $L_m$, we use a weak prior on the 3D shape of the object.
Specifically, we assume that a 3D template surface $S_0\subset\mathbb{R}^3$ is available, for example as simplicial (triangulated) mesh.
We fit the 3D surface to the LiDAR points by considering a rigid motion $g \in SE(3)$ which, applied to $S_0$, results in the posed mesh
$
S = gS_0 = \{g\X : \X \in S_0\}.
$
We then define a distance between the mesh and the 3D LiDAR points as follows:
\begin{align}\label{e:dist1}
d(S|L_m)
&=
\frac{1}{|L_m|}
\sum_{\X \in L_m}
\min_{\X' \in S}
\|\X' - \X \|^2.
\end{align}
This quantity is similar to a Chamfer distance, but it only considers half of it:
this is because most of the 3D points that belong to the template object are \emph{not} be visible in a given view (in particular, at least half are self-occluded), so not all points in the template mesh have a corresponding LiDAR point.


Given a 2D object mask $m$ and its corresponding LiDAR points $L_m$, we can find the pose $g$ of the object by minimizing $d(g S_0|L_m)$ with respect to $g \in SE(3)$.
Then the bounding box of the object $m$ is given by $gB_0$ where $B_0 \subset \mathbb{R}^3$ is the 3D bounding box that tightly encloses the template $S_0$.

In accordance with prior work~\cite{geiger12are-we-ready,qin20weakly,meng2020ws3d}, we can in practice carry out the minimization not over the of full space $SE(3)$, but only on 4-DoF transformation $g = [R_\theta,~\mathbf{T}]$ where the rotation $R_\theta$ is restricted to the \emph{yaw} $\theta$ (rotation perpendicular to the ground plane).
Even so, direct minimization of~\cref{e:dist1} is in practice prone to failure because individual partial LiDAR point clouds do not contain sufficient information and fitting results are thus ambiguous (we do not report results in this setting as they are extremely poor).



Our solution to the ambiguity of fitting~\cref{e:dist1} is to \emph{share information} across all object instances in the dataset.
We do this by training a deep neural network
$
\Phi : \operatorname{Fin}(\mathbb{R}^3) \rightarrow SE(3)
$
mapping the LiDAR points $L_m$ to the corresponding object pose $g = \Phi(L_m)$ directly.
The network $\Phi$ can be trained in a self-supervised manner by minimizing \cref{e:dist1} averaged over the entire dataset
$
\mathcal{D}
$
as
\begin{align}\label{e:simpleloss}
  \mathcal{L}(\Phi|\mathcal{D})
  &=
  \frac{1}{|\mathcal{D}|}
  \sum_{(L_m,m)\in\mathcal{D}}
  d(g_m S_0 | L_{m}),
  ~~~
  \text{where} ~ g_m = \Phi(L_m).
\end{align}

\subsection{Modelling and discounting outliers}

\begin{figure}
    \centering

        \begin{tabular}{c c c c}
            \includegraphics[width=0.25\textwidth]{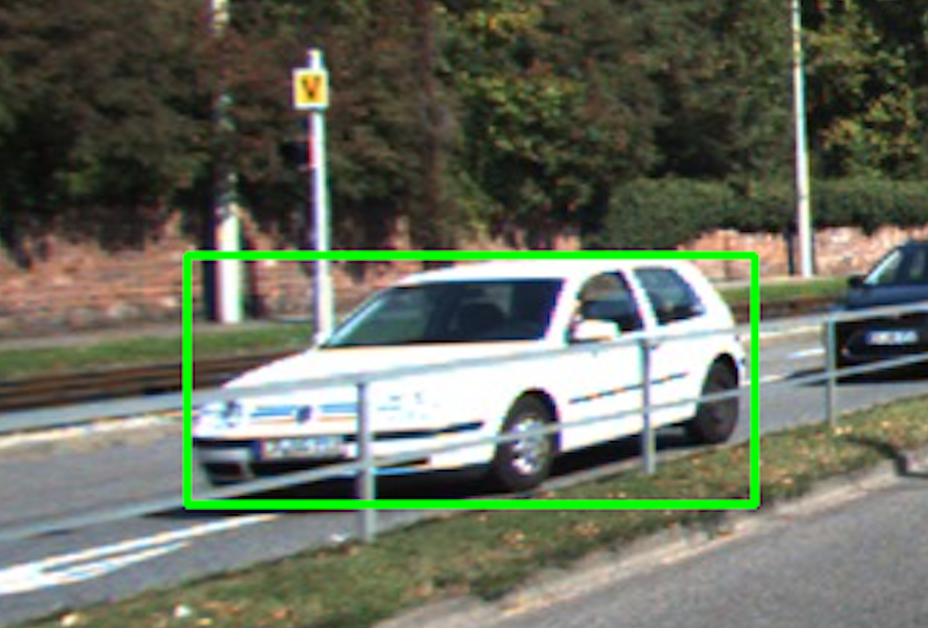} &
            \includegraphics[width=0.25\textwidth]{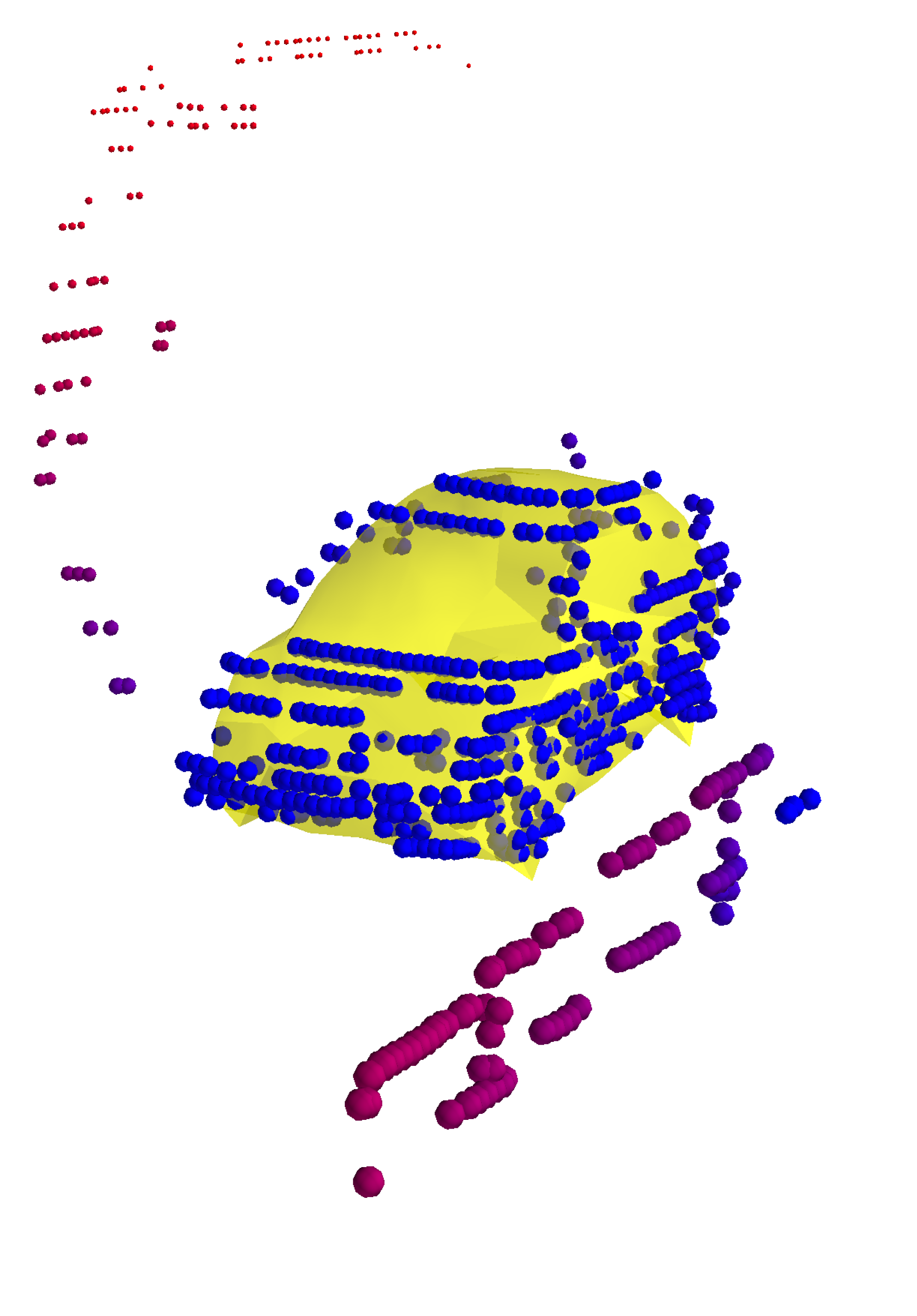} &
        \end{tabular}
        \caption{Even after removing LiDAR points which do not project into the 2D instance mask we still have a large number of outliers, mainly caused by partial occlusions and points at the boundary where 2D masks struggle. The $\sigma^2(\X)$ value predicts the relevance of a point to the cars location with blue points having low values with a gradient to red for higher values which suggests the point is an outlier.}
    \label{fig:my_label}
\vspace{-8mm}
\end{figure}

A major drawback of~\cref{e:simpleloss} is that LiDAR points tend to be noisy, especially because the boundaries of the region $m$ may not correspond to the object exactly or the LiDAR may be affected by a reflection or `see through' a glass surface.
Such points might disproportionally skew the loss term, forcing the estimated object position closer to these outliers.
In order to help the model discriminate between inliers and outliers, we let the network predict an estimate of whether a given measurement is likely to belong to the object or not.

We therefore propose a second network $\sigma(\X) = \Psi_\X(L_m)$ that assigns to each LiDAR point $\X \in L_m$ a variance $\sigma$ and jointly optimize $\Phi$ and $\Psi$ by minimizing~\cite{novotny17learning,kendall17what}:
\begin{equation}\label{e:distance}
  \bar d(S|L_m)
  =
  \frac{1}{|L_m|}
  \sum_{\X \in L_m}
  \min_{\X' \in \hat S}
  \frac{\|\X' - \X \|^2}{\sigma^2(\X)}
  +
  \log \sigma^2(\X)
\end{equation}
Note that the network $\Psi$ has to make a judgement call for every point $\X$ on whether it is likely to be an outlier or not \emph{without} having access to the loss.
A perfect prediction (i.e., one that minimizes the loss) would set $\sigma(\X) = \|\X' - \X \|$ to be the same as the fitting error.
The desirable side effect is that, in this manner, outliers are discounted by a large $\sigma(\X)$ when it comes to estimating the pose $g_m$ of the object.


\subsection{Direct optimization for the yaw}\label{s:direct-yaw}

Finally, we note that fitting the rotation of the object can be ambiguous, especially if only a small fragment of the object is visible in the RGB/LiDAR data.
Specifically, the distance $\bar d([R_{\theta_m}~\mathbf{T}_m]S_0|L_m)$, which is usually well behaved for the translation component $\mathbf{T}_m$, has instead a number of `deep' local minima, which we found is mainly caused by the inherent ambiguities of fitting the rotation $\mathbf{\theta}_m$ (yaw) parameter.
Each minimum corresponds to a $90^{\circ}$ rotation as in many example only a single side of the vehicle is visible.
In practice, as we show, the yaw network $\Psi$ \emph{can} learn to disambiguate the prediction, but it usually fails to converge to such a desirable solution without changes in the formulation.

In order to solve this issue, we propose to modify the formulation to incorporate \emph{direct optimisation} over the yaw.
In other words, every time the loss is evaluated, we assess a number of possible rotations $R$, as follows:
%
%
\begin{align}
  R^*_m
  &= \operatornamewithlimits{argmin}_{R\in \mathcal{R}}
  \bar d([R ~~\mathbf{T}_m] S_0 | L_{m})
  \\\label{e:iteratedloss}
  \mathcal{L}'(\Phi, \Psi|\mathcal{D})
  &=
  \frac{1}{|\mathcal{D}|}
  \sum_{(L_m,m)\in\mathcal{D}}
  \bar d([R^*_m~ \mathbf{T}_m] S_0 | L_{m}) + \lVert R_m - R_m^*\rVert .
\end{align}

The loss is the same as before for the translation, but, given the predicted translation, it always explores all possible rotations $R$, picking the best one $R_m^*$.

Note that this does not mean that the network $\Phi$ is not tasked with predicting a rotation anymore; on the contrary, the network is encouraged to output the optimal $R_m^*$ via minimization of the term $\|R_m - R_m^*\|$.
This has the obvious benefit of not incurring the search at test time, and therefore the final network running time is not affected by this process.

In practice, as only the yaw angle is predicted, we implement this loss by quantizing the interval $[0, 2\pi)$ in 64 distinct values (bins), as in our experiments we found this number of bins sufficient (see results in \cref{t:yaw}).
In this manner, the rotation head of the network $\Phi$ can be interpreted as a softmax distribution $\Phi_r(L_m)$ and the norm $\lVert . \rVert$ is replaced by the cross-entropy loss.

\subsection{Multi-frame consistency}

\begin{figure}
    \centering
    \begin{tabular}{c c}
        \multicolumn{2}{c}{Frame 1} \\
        \includegraphics[height=0.16\textwidth]{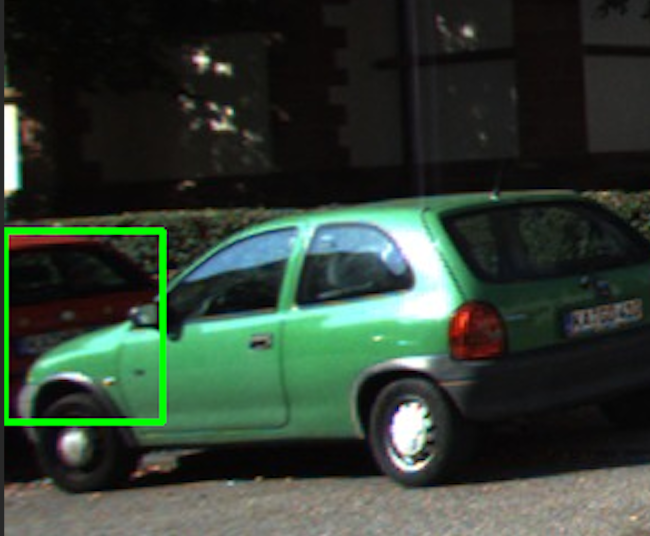} &
        \includegraphics[width=0.15\textwidth]{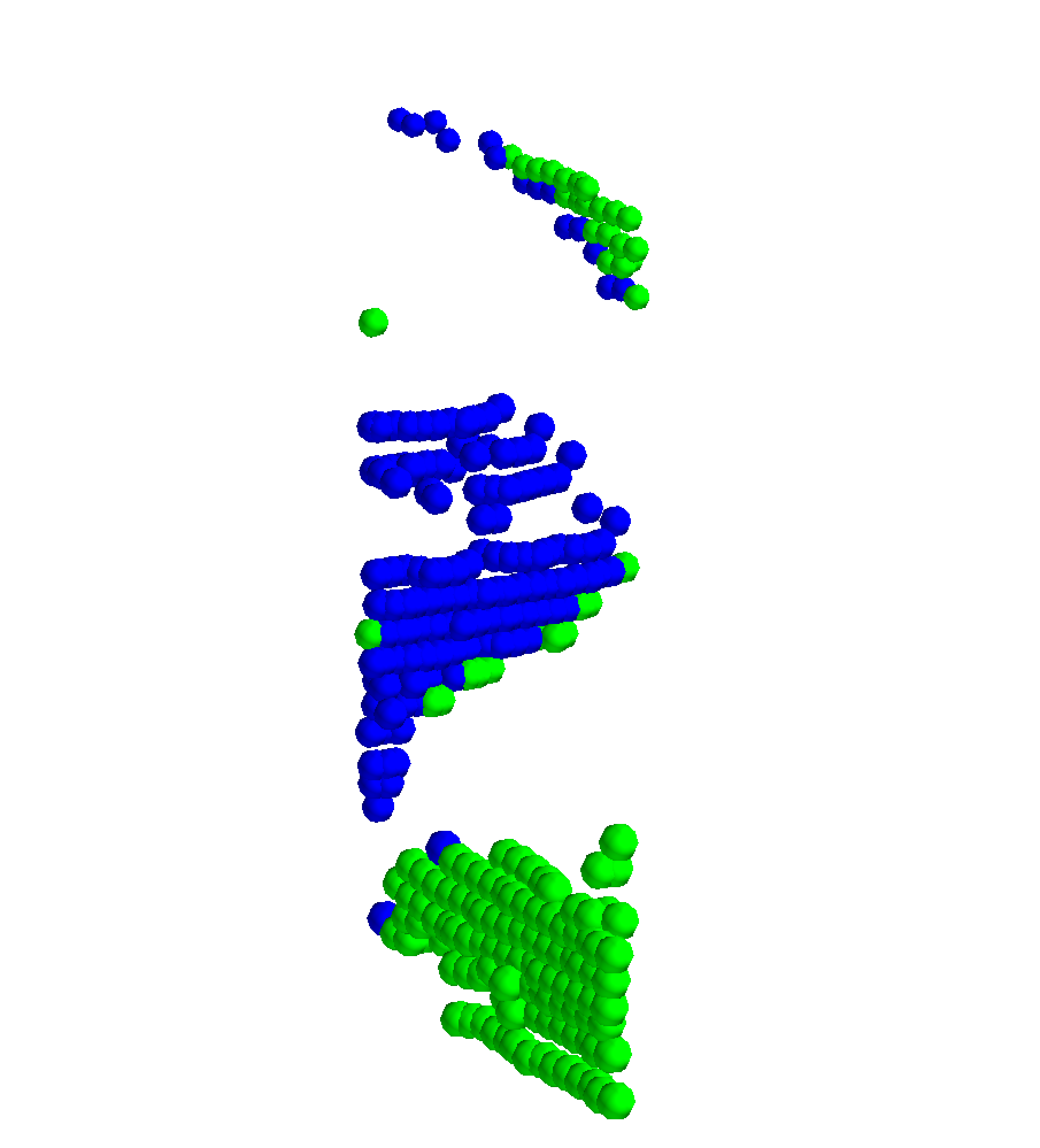} \\
        \multicolumn{2}{c}{Frame 5} \\ 
        
        \includegraphics[height=0.16\textwidth]{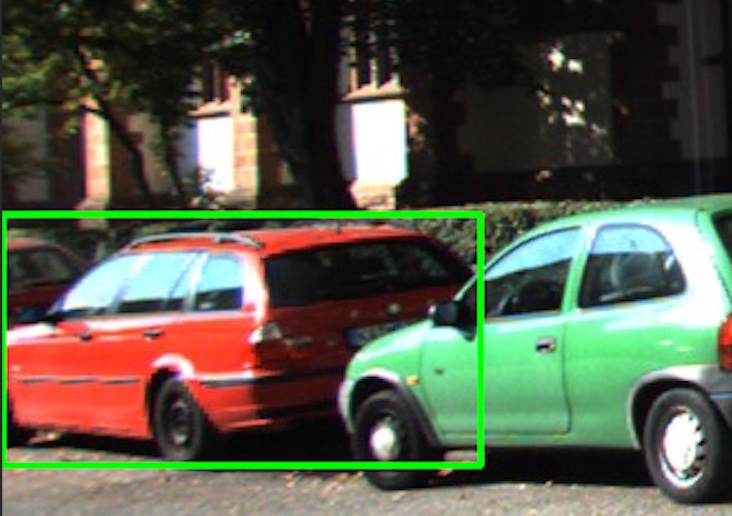} &
        \includegraphics[height=0.2\textwidth]{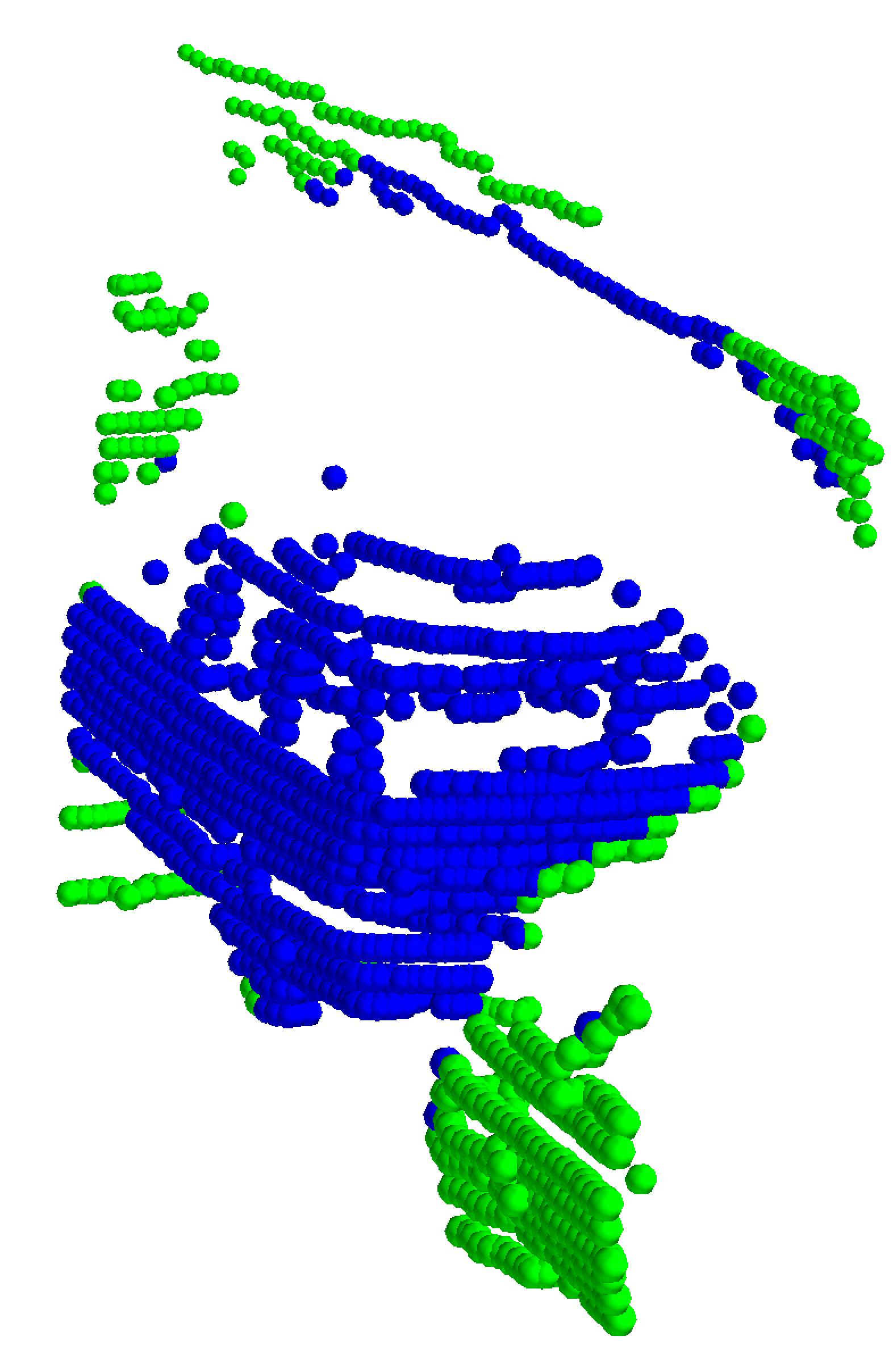} \\
        
    \end{tabular}
    \caption{Two frames in a sequence of five, with a heavily occluded and truncated car in the first frame and a much better view of the same in the last.
    Multi-frame consistency allows the method to use clearer frames to interpret more difficult ones, which is particularly helpful in discovering the outliers.}\label{fig:multiframe_example}
\vspace{-8mm}
\end{figure}

Whilst in our approach we do not have the actual 3D pose of the object available as a training signal, the 3D pose of the object across multiple frames must be consistent with the observer ego-motion.
This is true for vehicles that are not moving (parked cars) and approximately true for other vehicles; in particular, the \emph{yaw} of the objects, once ego-motion has been compensated for, is roughly constant.





Specifically, consider an arbitrary model point $\X_0\in S_0$.
Observed in a LiDAR frame $i$, this point is estimated to be at location $\X_0^i = g^i \X_0$, where $g^i$ is the network prediction for frame $i$.
Likewise, let $\X_0^j = g^j \X_0$ be the point position at frame $j$.
If the object is at rest, the two positions are related by
$
    \X^i_0 = g_{i \leftarrow j} \X^j_0
$
where $g_{i \leftarrow j}$ is the ego-motion between frames $i$ and $j$ of the vehicle where the sensors are mounted, which we assume to be known.
We can thus define the consistency loss $\mathcal{L}^{\X_0}$ for any model point $\X_0$:
\begin{equation}\label{e:keypointloss}
   \mathcal{L}^{\X_0}(\Phi|\mathcal{D})
=
  \frac{1}{|\mathcal{D}|}
  \sum^N_{i=1}
  \sum_{(L_m,m)\in\mathcal{D}^i}
  \sum^K_{j=1} \lVert\ g_{i \leftarrow i + j}
  \X_0^{i+j} - \X_0^i
  \rVert^2.
\end{equation}
where $\mathcal{D}^i$ denotes LiDAR-mask pairs detected in the frame $i$, $K$ is the length of the frame sequence over which the consistency is evaluated, and $N$ is the total number of frames.

\begin{figure}
    \centering
    \begin{tabular}{c c}
        \includegraphics[width=0.5\textwidth]{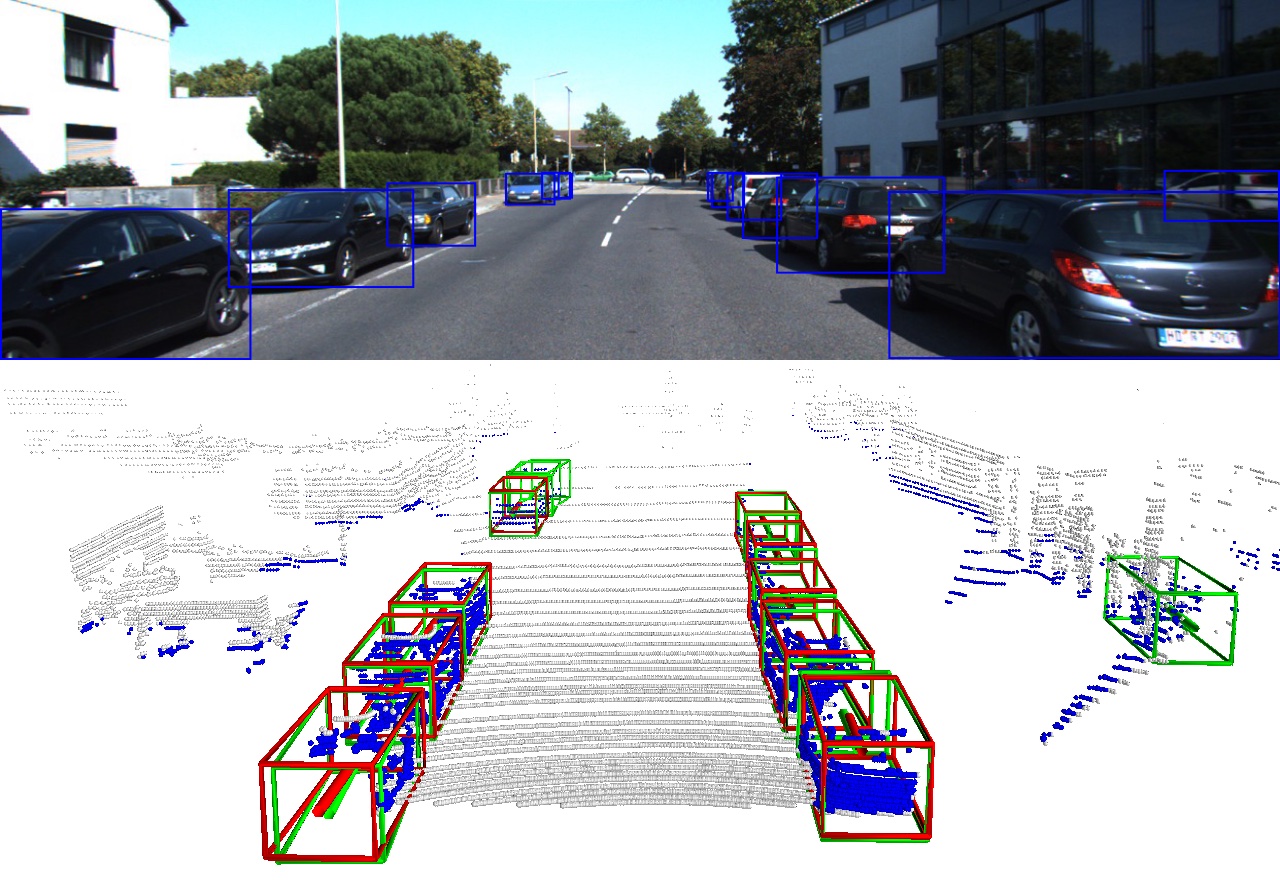} 
         
        
        \vspace{-5pt}

    \end{tabular}
    \caption{Qualitative results of our method (green) on the KITTI Validation Set with ground truth car annotations in red and points inside each mask in blue.\vspace{-5pt}}
    \label{f:Qualitative}
\vspace{-4mm}
\end{figure}

Note that loss~\eqref{e:keypointloss} is null for all object keypoints if, and only if, $g^i = g_{i\leftarrow j}g^j$, which is an \emph{equivariance} condition for the predictor.
However, we found it more convenient and robust to enforce this equivariance by using~\cref{e:keypointloss} summed over a small set of representative model points.
In our experiments we have found that at least two points are required to ensure that we consistently predict the heading (yaw) of the detected car.
In practice, we use car centre and front keypoints (see \cref{f:Qualitative}), so the final loss term used to train the model is:
\begin{equation}\label{e:finalloss}
\mathcal{L}(\Phi, \Psi|\mathcal{D}) =
\mathcal{L}'(\Phi,\Psi|\mathcal{D})
+ \mathcal{L}^{\X_\text{centre}}(\Phi|\mathcal{D})
+ \mathcal{L}^{\X_\text{front}}(\Phi|\mathcal{D})
\end{equation}

\paragraph{Implementation details of tracking}

A detector such as Mask R-CNN only provides 2D mask for individual video frames, but defining the loss~\eqref{e:keypointloss} requires to identify or track the same object across two frames.
For tracking, we take the median of LiDAR points for each mask and compare them to the medians of masks in adjacent frames.
The closest pseudo-centres are chosen to be the same vehicle if and only if the number of LiDAR points has not changed significantly and the distance between them is less than 2 meters when ego motion is accounted for.
The distance criterion ensures that the same vehicle is detected while the number of points removes poor Mask-RCNN detections in subsequent frames. This tracking is only used at training time, at test time we use all 2D detections.

\section{Experiments}\label{s:experiments}

We assess our methods against the relevant state of the art on standard benchmarks.

\paragraph{Data}\label{s:expsetup}

For our experiments, we use the KITTI Object Detection dataset~\cite{geiger12are-we-ready} which has 7481 frames with labels in 3D.
However, we do so for compatibility with prior work.
Specifically, we use the split found in~\cite{chen2017multiview}, the standard across all prior works which  first splits the videos into training and validation sets focusing on two different parts of the world (no visual overlap).
The network is learned on the training videos using multiple frames and then applied and evaluated on the validation videos on a single-frame basis.

KITTI evaluates vehicle detectors using Birds Eye View (BEV) IOU and 3D box IOU (3D) with a strict cutoff of $0.7$ for a positive detection.
To be compatible with other relevant works in automated labelling~\cite{sdflabel, qin20weakly}, we evaluate instead at a threshold of $0.5$, also used for other KITTI categories, which reduces the influence of object size on IoU performance.
In part, this is motivated by~\cite{feng2020labels} which notes that the size of the `ground-truth' KITTI annotations are often imprecise due to the fact that many object instances have few LiDAR points and thus it is difficult for human annotators to accurately estimate metric size.


\paragraph{Data preprocessing}

To construct our training set we first run Mask-RCNN~\cite{he17mask,wu2019detectron2} with the ResNet 101 backend to locate cars in the images.
We then extract the LiDAR points contained within the masks detected and use these for 3D labelling.
When tracking for 5 frames this gives us 9.6k cars in the training.
For evaluation we only use a single frame of Mask R-CNN detection and use the mask score as the confidence for our predictions.

\paragraph{Implementation}

We implement our method in Pytorch \cite{NEURIPS2019_9015} and use components from Frustum PointNets \cite{qi2017frustum} to construct our network.
All variants are trained with Adam optimizer with a learning rate of $3\times 10^{-3}$ decreasing every 30 steps with multiplier $0.3$ for a total of 150 epochs with a batch size of 64 .
Training was performed on a single Titan RTX with a Ryzen 3900X processor and the most complex models had a training time of 7 hours.


\begin{table}
\setlength{\tabcolsep}{1.5mm}
\centering
\caption{Ablation of different components of the model and data processing steps. Our full model with automatic outlier removal and multi-view consistency achieves the highest accuracy.}\label{t:ablation}
\begin{tabular}{c c c c c c c c c c}
\toprule
    & \multicolumn{3}{c}{Components} & \multicolumn{3}{c}{AP\textsubscript{BEV}(IoU = 0.5)} \\
    & Filtering                      & Outlier-aware                                        & Multi-view                                            & Easy           & Moderate       & Hard           \\
 \cmidrule(r){2-4}
 \cmidrule(lr){5-7}
 \cmidrule(l){8-10}
(a) & 2D Box                         &                                                      &                                                       & 35.53          & 41.54          & 33.96          \\
(b) & \cellcolor{green}2D Mask       &                                                      &                                                       & 58.61          & 62.56          & 54.20          \\
(c) & \cellcolor{green}2D Mask       &                                                      & \cellcolor{green}\checkmark                           & 58.63          & 61.70          & 54.11          \\
(d) & \cellcolor{green}2D Mask       & \cellcolor{green}\checkmark                          &                                                       & 75.46          & 76.60          & 68.59          \\
(e) & \cellcolor{green}2D Mask       & \cellcolor{green}\checkmark                          & \cellcolor{green}\checkmark                           & \textbf{80.73} & \textbf{81.70} & \textbf{73.61} \\
\bottomrule
\end{tabular}
\vspace{-4mm}
\end{table}

\begin{table}
\centering
\caption{Comparison of yaw prediction techniques.}\label{t:yaw}
\begin{tabular}{c l c c c  c c c  }
\toprule
   &\multirow{2}{*}{Paradigm}                                 & \multicolumn{3}{c}{AP\textsubscript{BEV}(IoU = 0.5)} \\
   &                                                                & Easy      & Moderate      & Hard      \\
\cmidrule(lr){2-2}
\cmidrule(lr){3-5}
\cmidrule(lr){6-8}
(a)& arctan                                                         & 76.65                 & 79.05             & 69.33         \\
(b)&Insafutdinov \& Dosovitskiy~\cite{insafutdinov18pointclouds}    & 77.28                 & 76.92             & 69.12         \\
(c)&Goel et al.~\cite{goel20shape}                                  & 76.49                 & 77.35             & 69.69         \\
\midrule
(d)&Ours, 16 bins                                                   & 77.04                & 77.30              & 69.49         \\
(e)&Ours, 32 bins                                                   & 79.93                & 81.14              & 73.15         \\
(f)&Ours, 64 bins                                                   & \textbf{80.73}       & \textbf{81.70}     & \textbf{73.61} \\
(g)&Ours, 128 bins                                                  & 81.79                & 82.23              & 74.11         \\
\bottomrule
\end{tabular}

\vspace{-4mm}
\end{table}

\subsection{Ablations of model components}

We first start by analysing the individual components of our model (\cref{t:ablation}).
First, we do not use the 2D mask to filter LiDAR points, significantly increasing the number of outliers, arising in particular from cars proximal to the target one (row (a)).
Masking LiDAR points (row (b)) results in a substantial improvement, removing most of these outliers (see also \cref{fig:dataset_example}).
Introducing the temporal consistency/equivariance loss~\eqref{e:keypointloss} (row (c)) does not give by itself a noticeable benefit because outlier points are still heavily influential to the prediction.
Discounting outliers using the probabilistic formulation of \cref{e:distance} increases accuracy substantially (row (d)).
Furthermore, bringing back the consistency loss, which amounts to our full model, does now show a significant benefit (row (e)).
Our interpretation is that considering multiple frames can significantly aid discovering and learning outlier patterns:
this is because outliers tend to be \emph{inconsistent} across frames, so reasoning over multiple frames helps discovering them (see \cref{fig:multiframe_example}).

\paragraph{Yaw estimation}

In \cref{t:yaw} we evaluate our approach for estimating camera viewpoint proposed in \cref{s:direct-yaw}.
First we experiment with a network $\Phi$ tasked to output a vector $\mathbf{x} \in \mathbb{R}^2$ with the yaw angle computed as $\theta = \arctan(x_1/x_2)$ (row (a)).
Our direct prediction approach (rows (d-g)) outperforms this na\"ive baseline by a significant margin.
Our method is influenced by the number of discrete rotations, 64 being optimal (row (f)).
In order to provide stronger baselines, we additionally implement to alternative techniques~\cite{insafutdinov18pointclouds,goel20shape} to handle ambiguous predictions in 3D pose estimation, but did not observe a benefit (rows (b) and (c)) compared to simple direct arctan regression.
This is perhaps due to the different setting (\cite{insafutdinov18pointclouds,goel20shape} were proposed to handle ambiguous fitting of 3D shapes to 2D silhouettes).



\paragraph{SoTA comparison}
\begin{table*}[b]
\newcommand{\zero}{ZERO}
\centering

\caption{Object Detection Average-Precision on the KITTI validation set. Compared to our Method VS3D\cite{meng2020ws3d} uses a network trained on Pascal 3D and NYC 3D Cars\cite{xiang_wacv14, MatzenICCV13} to determine the object Yaw and 2D box, while while Zakharov\cite{sdflabel} only considers MASK R-CNN detections with an IOU > $50\%$ compared to a ground truth box and uses a synthetic dataset to train a network which gives yaw.
}\label{t:averageprecision}
\setlength{\tabcolsep}{1mm}
\begin{tabular}{cc c c c  c c c}
\toprule
&\multirow{3}{*}{Method} & \multicolumn{3}{c}{Annotations Source} & \multicolumn{3}{c}{AP\textsubscript{BEV}~/~AP\textsubscript{3D} (IoU = 0.5)} \\
&& \textbf{2D} & \multicolumn{2}{c}{\textbf{3D}}  \\
&& Boxes & Yaw & Boxes & Easy & Moderate & Hard \\
\cmidrule(r){1-2}
\cmidrule(lr){3-3}
\cmidrule(lr){4-5}
\cmidrule(l){6-8}
(a) &VS3D~\cite{meng2020ws3d} & Pascal 3D & NYC Cars & & 74.5/40.32 & 66.71/37.36 & 57.55/31.09 \\
\hline
(b) &Zakharov et al. \cite{sdflabel} & KITTI & KITTI &  & 77.84/62.25 & 59.75/42.23 & -/- \\
\hline
(c) &\textbf{Ours} &MS-COCO & &  & 80.73/76.73 & 81.70/76.66 & 73.61/69.01 \\
(d) & \textbf{Ours} & Cityscapes & & & \textbf{86.52/83.45} &	\textbf{86.22/79.53}	& \textbf{75.53/71.01} \\
\hline
(e) &\textit{Frus.PointNet~\cite{qi2017frustum} }     & \textit{KITTI} & \textit{KITTI} & \textit{KITTI} & \textit{98.16/97.67}     & \textit{94.80/93.81}     & \textit{87.11/86.14} \\

\bottomrule
\end{tabular}

\vspace{-8mm}
\end{table*}
In \cref{t:averageprecision} we compare our method to the relevant state of the art.
VS3D~\cite{qin20weakly} uses a viewpoint estimation network pretrained on Pascal 3D and NYC 3D cars~\cite{xiang_wacv14, MatzenICCV13} with ground truth yaw annotations.
In Zakharov et al.~\cite{zakharov20autolabeling} a synthetic dataset is used to initialise a coordinate shape space NOCS~\cite{Wang_2019_CVPR} providing a strong prior on 3D shape and yaw estimation.
During fitting stage they only utilise Mask R-CNN predictions which have a high IOU compared to a ground truth 2D box and also takes 6 seconds to infer a single car on a modern GPU making it infeasible for real time prediction unlike VS3D~\cite{qin20weakly} (22Hz) and our method which runs at 200Hz after the 2D object detection method (about 25Hz). We provide results (c) with a 2D detector trained on MS-COCO\cite{lin14microsoft} for fairest comparison to this method which has pretrained requirements closest to our own.
Frustum PointNet~\cite{qi2017frustum} is a fully supervised (2D box, yaw, 3D size, 3D centre) method trained on KITTI that serves as a reference with similar architecture to our method.




\paragraph{Auto-Label Generation}
In \cref{t:autolabel} we compare different supervision options on a single model. We train the FPN~\cite{qi2017frustum} using original 3D ground truth, as well as only the 3D ground truth where there is a matching 2D detection from Mask-RCNN~\cite{wu2019detectron2} (the same detector used in our method) to demonstrate what is the impact of instances missed by the 2D detector to the overall performance. We show that the difference between our labelling method, which does not use any 3D ground truth, and using full 3D ground truth of only the instances that are ``made available'' by the 2D detector, is relatively small, and therefore the 2D detection quality is the main limiting factor.


\begin{table}[H]
\centering
\caption{Using different forms of supervision to train Frustum PointNet~\cite{qi2017frustum}}\label{t:autolabel}
\begin{tabular}{c c c c }
\toprule
Supervision & \multicolumn{3}{c}{AP\textsubscript{BEV}(IoU = 0.5)} \\
 & Easy           & Moderate       & Hard \\
\hline
   
Fully Supervised (original GT) & 98.16 & 94.80 & 87.11 \\

Fully Supervised (GT w/ matching dets.) & 91.83 & 87.16 & 78.29 \\
\textit{Supervised by our labels (no GT)} & 90.23 & 85.74 & 76.84 \\

\bottomrule
\end{tabular}

\vspace{-4mm}
\end{table}

\section{Conclusions}\label{s:conclusions}

In this paper we presented a novel method of localising 3D objects in LiDAR point clouds, trained using only generic 2D object detector. Compared to previous work, our method is less complex, we do not require the use of additional manually annotated data sources as in \cite{qin20weakly} or virtual data and ground truth 2D bounding boxes as in \cite{zakharov20autolabeling}, and we still achieve superior accuracy and better run time. To our knowledge, our method is the first method that can learn to transform 2D predictions to 3D detections without any need for supervision of 3D parameters.

\newpage
{\small \bibliographystyle{plain}\bibliography{mccraith_g,vedaldi_general,vedaldi_specific, insafutdinov}}

\end{document}



\section{Additional Results}
In table \ref{t:averageprecision_supp} we provide results for our method with the Mask R-CNN detector trained on the more application-specific Cityscapes\cite{cordts16the-cityscapes} dataset. In particular, running Cityscapes-trained detector on KITTI validation set produces approximately $12000$ detections, whereas COCO-trained detector yields almost $15600$ detections, many of which are of a general vehicle class and not annotated as cars in KITTI. This has advantages at training time, as our network's capacity is not used to attempt fitting non-car objects and also at test time, as we get a lower false positive rate, which results in an overall higher performance.

\begin{figure}
    \centering
    \begin{tabular}{c c}
        
            \includegraphics[width=0.5\textwidth]{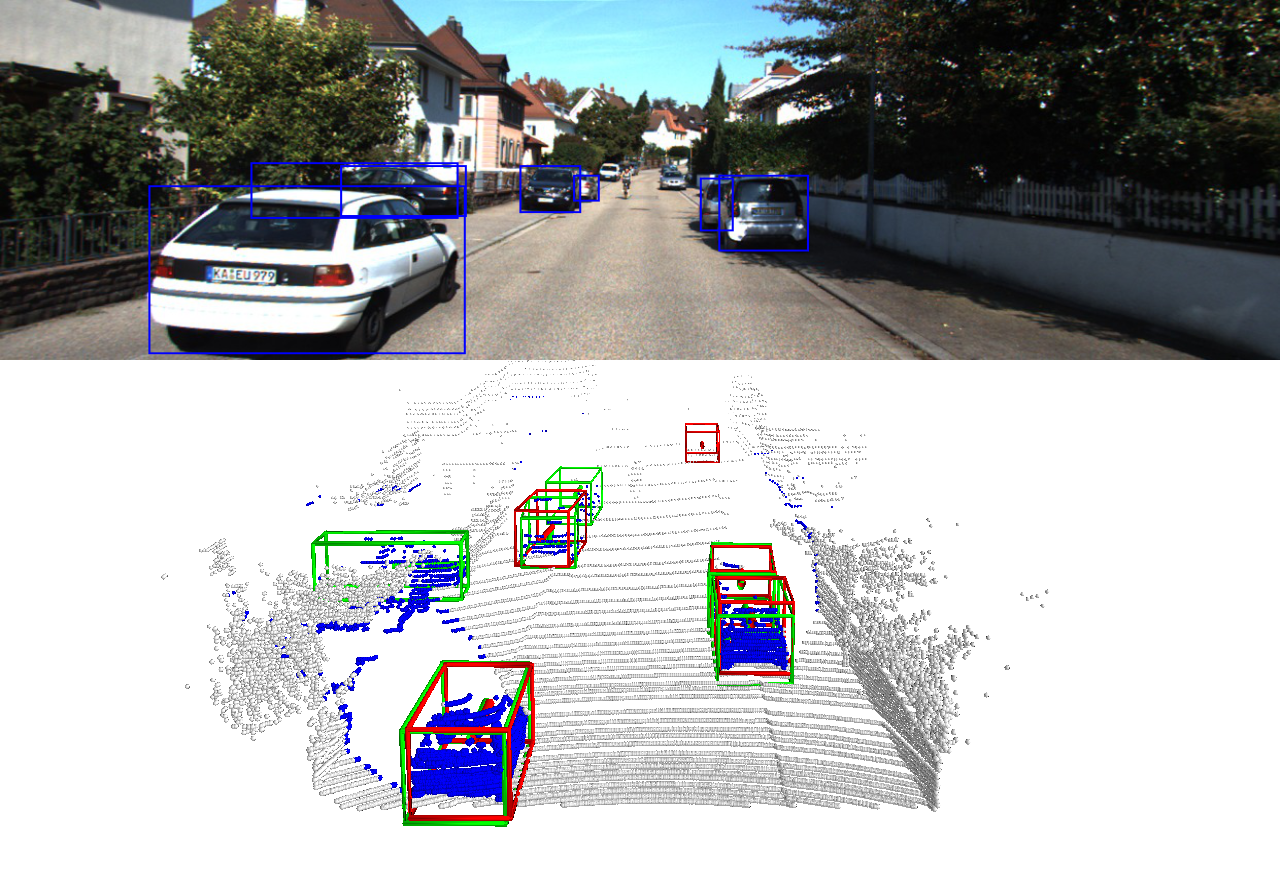}
            & 
            \includegraphics[width=0.5\textwidth]{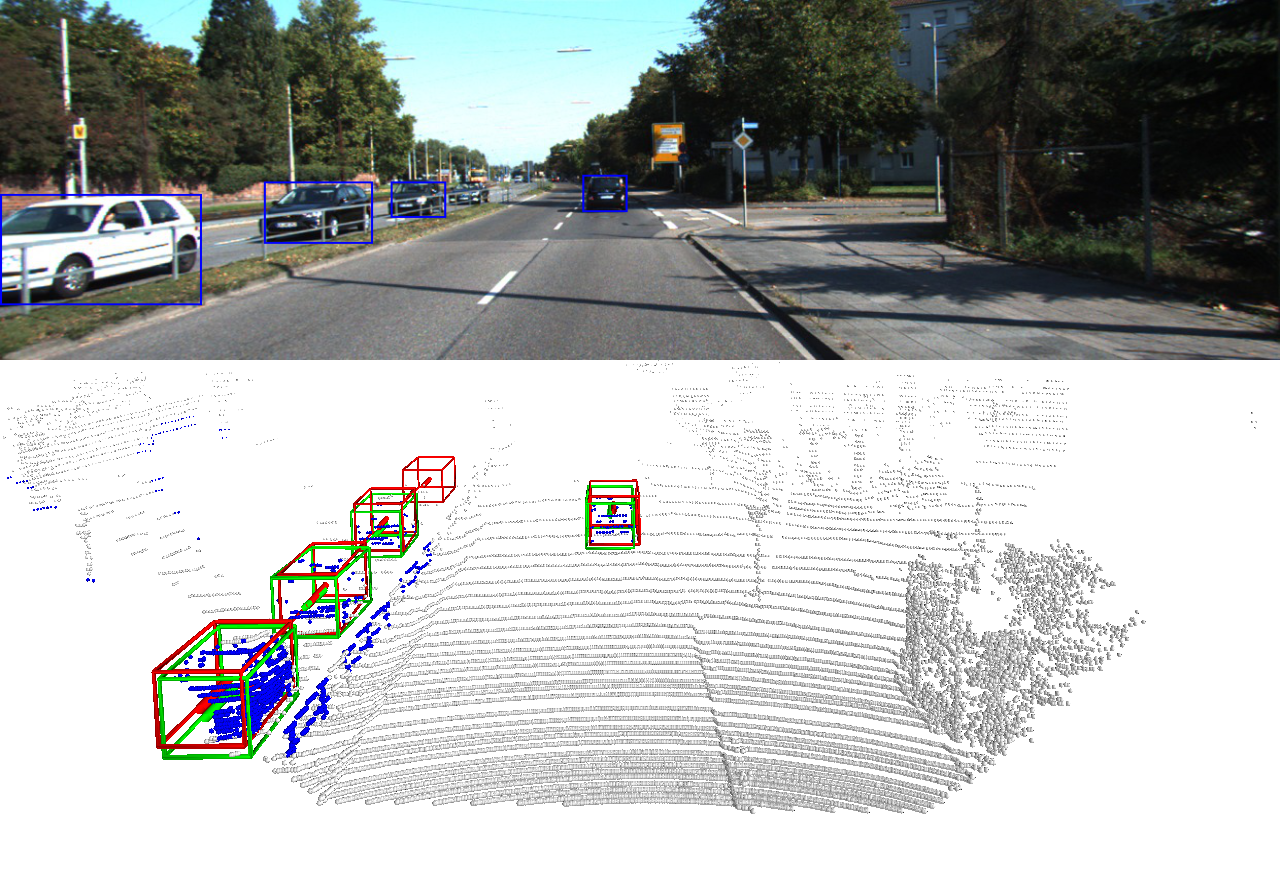}
            \\
            a & b\\
            \includegraphics[width=0.5\textwidth]{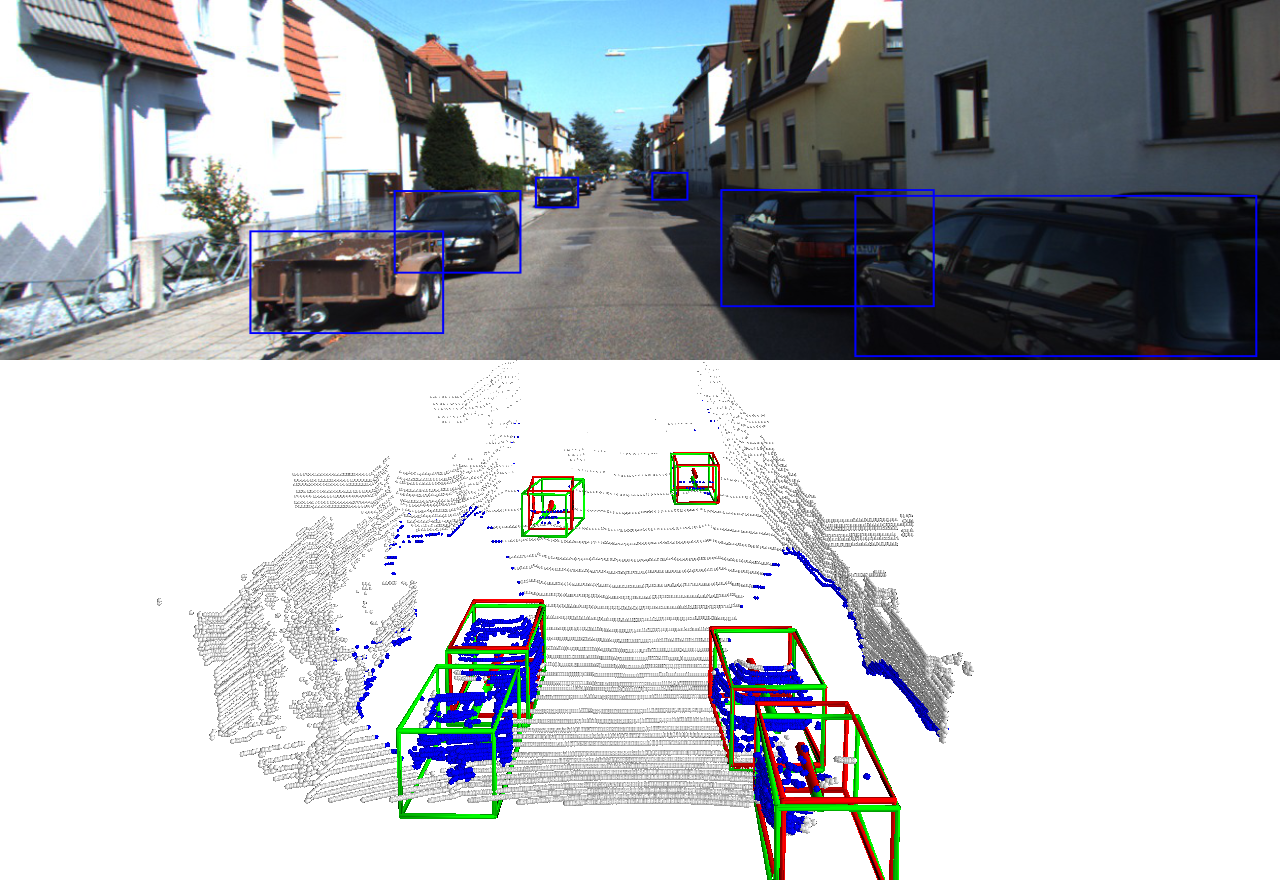}
            &  
            \includegraphics[width=0.5\textwidth]{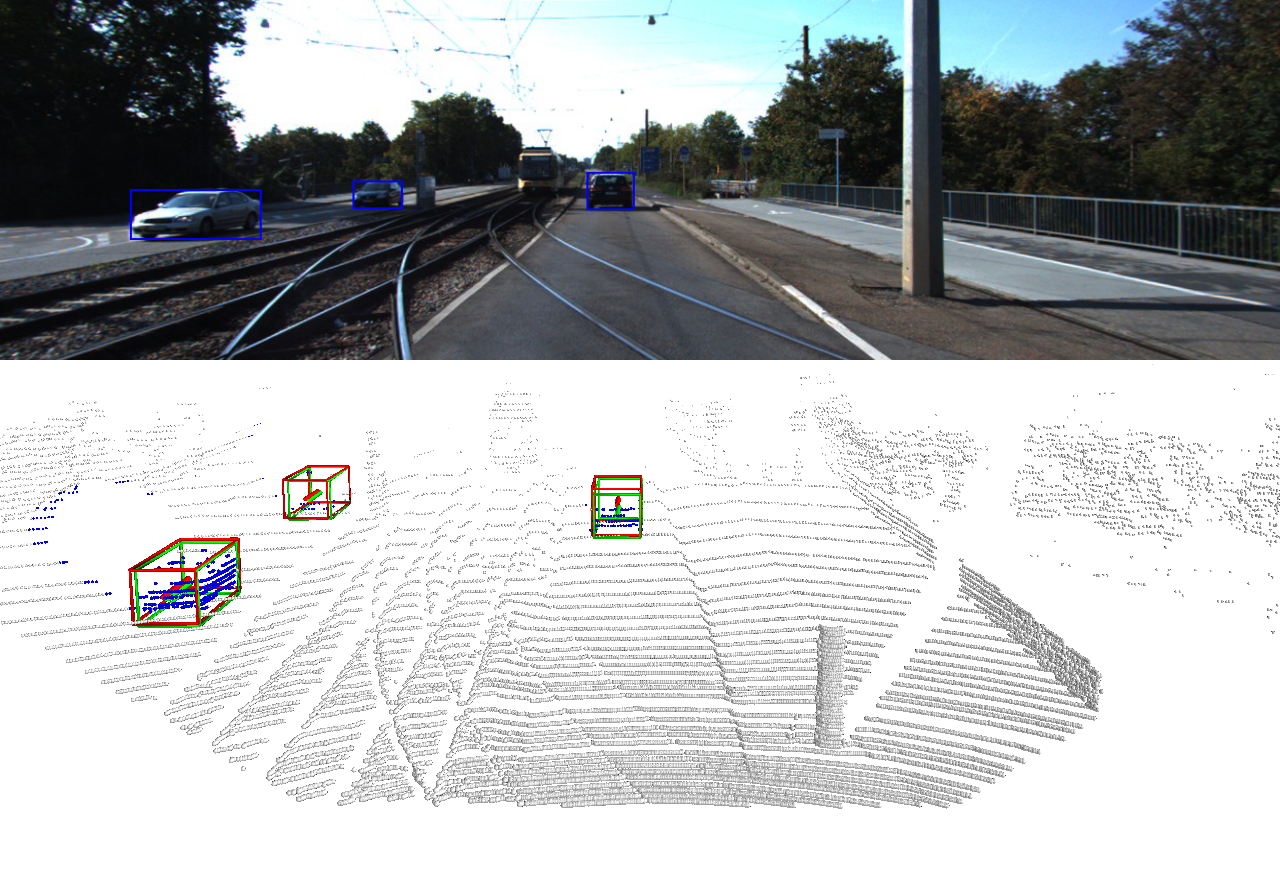}
            \\
            c & d\\
        
            \includegraphics[width=0.5\textwidth]{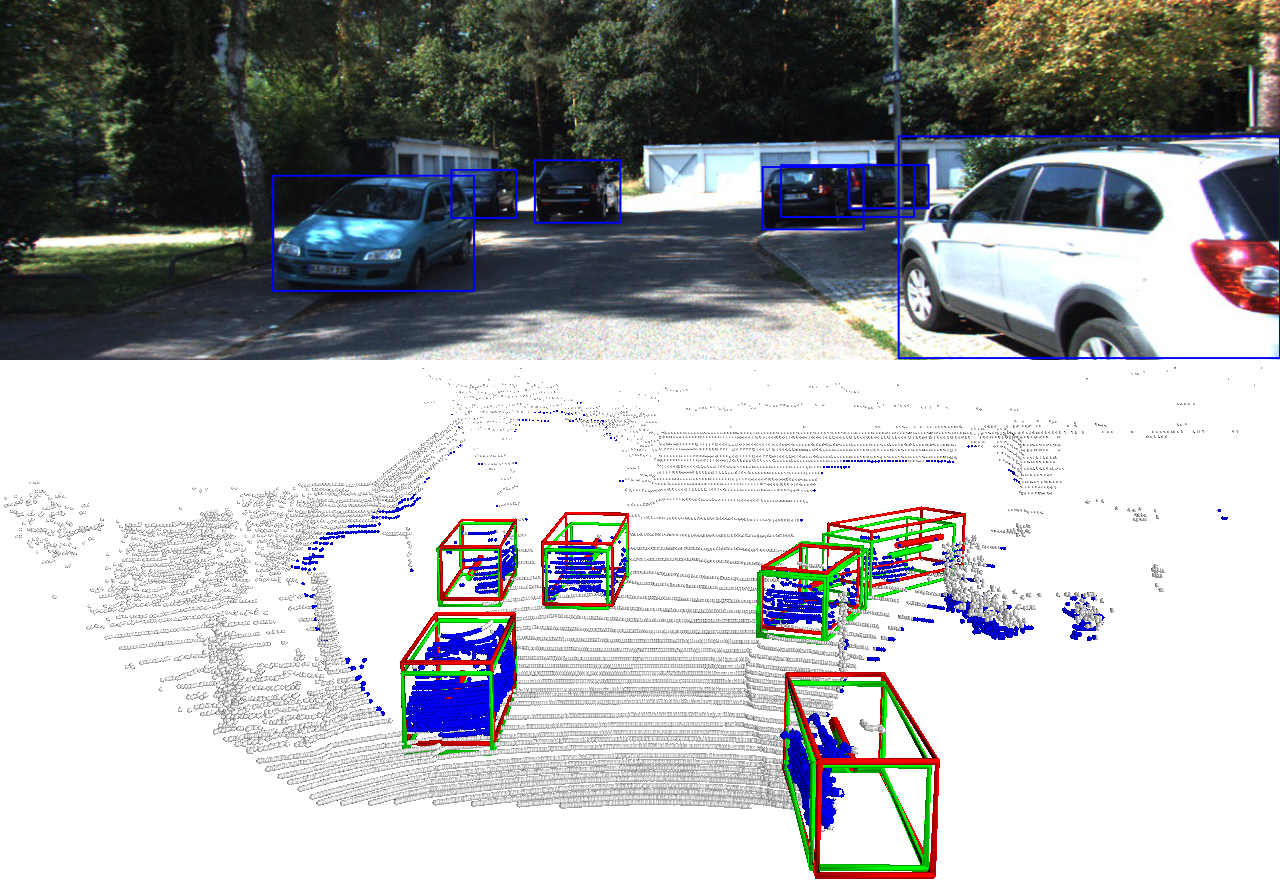}
            & 
            \includegraphics[width=0.5\textwidth]{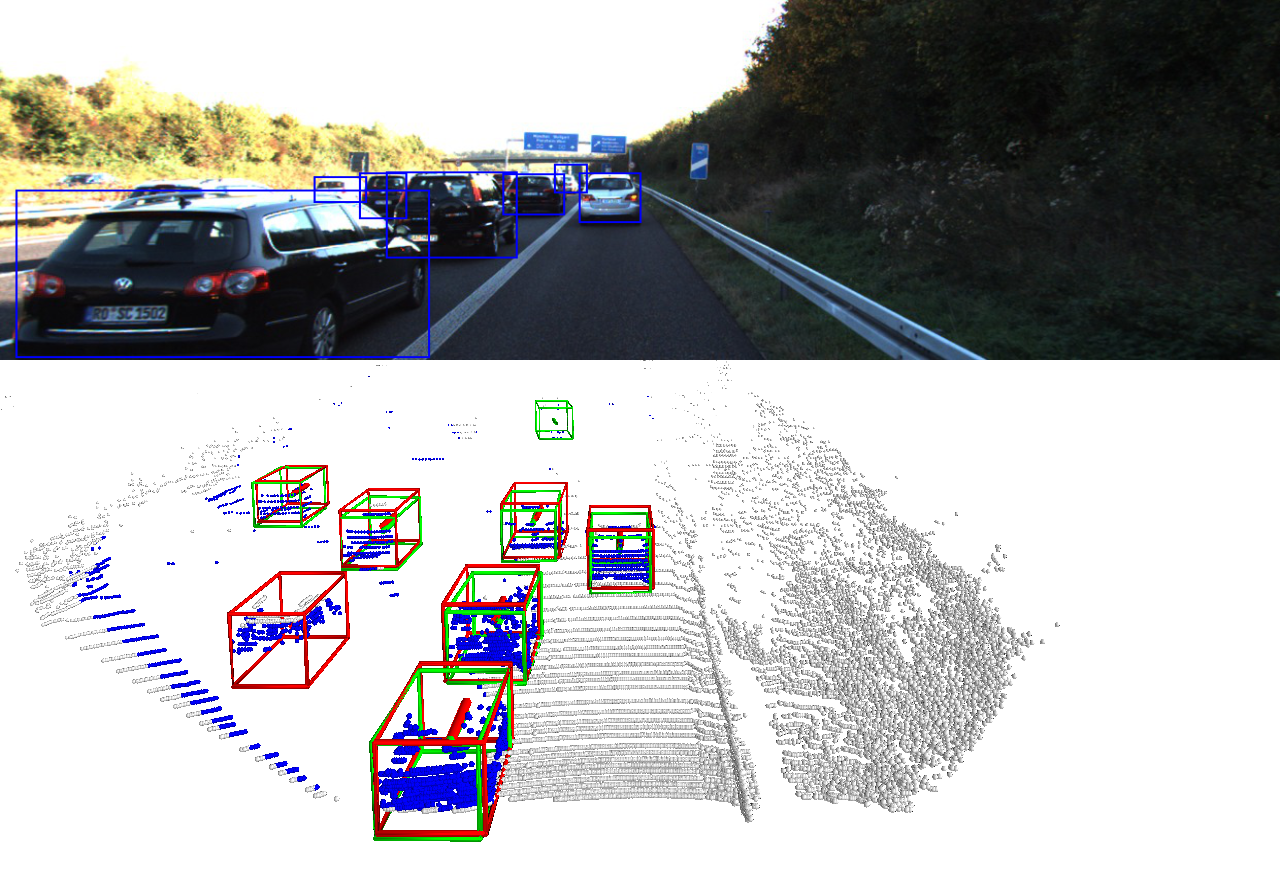}
            \\
            e & f\\

    \end{tabular}
    \caption{Ours in Green, Ground Truth in Red.}
    \label{f:Qualitative_supp}
\end{figure}

We also provide additional qualitative examples in figure \ref{f:Qualitative_supp}. In (a,i,l)e of these images we observe cars predicted in green which are not annotated in red, some of this can be attributed to KITTI's particular classification of an object to a non-car vehicle category. In other cases (l) we see cars annotated further away which are annotated in future frames (m), implying issues with KITTI not annotating every car all the time. Two images (c,j) also contain trailers which COCO trained Mask R-CNN detected as cars. KITTI annotates about 3000 examples for each difficulty category with some sequences having annotations frame by frame, some every second frame and other sequences not containing many annotations which may make some detection errors more prominent than others if the number of annotated frames where an object erroneously detected are denser.

\begin{table*}[t]
\newcommand{\zero}{ZERO}
\centering
\setlength{\tabcolsep}{1mm}
\begin{tabular}{cc c c c  c c c}
\toprule
&\multirow{3}{*}{Method} & \multicolumn{3}{c}{Annotations Source} & \multicolumn{3}{c}{AP\textsubscript{BEV}~/~AP\textsubscript{3D} (IoU = 0.5)} \\
&& \textbf{2D} & \multicolumn{2}{c}{\textbf{3D}}  \\
&& Boxes & Yaw & Boxes & Easy & Moderate & Hard \\
\cmidrule(r){1-2}
\cmidrule(lr){3-3}
\cmidrule(lr){4-5}
\cmidrule(l){6-8}

(a) &VS3D~\cite{meng2020ws3d} & Pascal 3D & NYC Cars & & 74.5/40.32 & 66.71/37.36 & 57.55/31.09 \\
\hline
(b) &Zakharov et al. \cite{sdflabel} & KITTI & KITTI &  & 77.84/62.25 & 59.75/42.23 & -/- \\
\hline
(c) &\textbf{ours} &MS-COCO & &  & 80.73/ 76.73 & 81.70/76.66 & 73.61/69.01 \\


(e) &\textbf{ours} & Cityscapes & &  & \textbf{86.51}/ \textbf{81.47} & \textbf{84.36}/\textbf{75.49} & \textbf{75.49}/\textbf{68.30} \\


\hline
(g) &\textit{Frus.PointNet~\cite{qi2017frustum} }     & \textit{KITTI} & \textit{KITTI} & \textit{KITTI} & \textit{98.25/98.10} & \textit{94.92/94.31} & \textit{87.14/86.48} \\
\bottomrule
\end{tabular}
\caption{Object Detection Average-Precision on the Kitti validation set. Compared to our Method VS3D\cite{meng2020ws3d} uses a network trained on Pascal 3D and NYC 3D Cars\cite{xiang_wacv14, MatzenICCV13} to determine the object Yaw and 2D box, while while Zakharov\cite{sdflabel} only considers MASK R-CNN detections with an IOU > $50\%$ compared to a ground truth box and uses a synthetic dataset to train a network which estimates orientation. Using Mask R-CNN trained on Cityscapes\cite{cordts16the-cityscapes} instead of the more generic MS-COCO dataset allows us to reduce the number of false positives. }\label{t:averageprecision_supp}
\end{table*}

\begin{figure}
    \centering
    \begin{tabular}{c c}

        \includegraphics[width=0.5\textwidth]{figures/Qualitative_examples/2.png} &
        \includegraphics[width=0.5\textwidth]{figures/Qualitative_examples/166.png} \\
        g & h \\ 
        \includegraphics[width=0.5\textwidth]{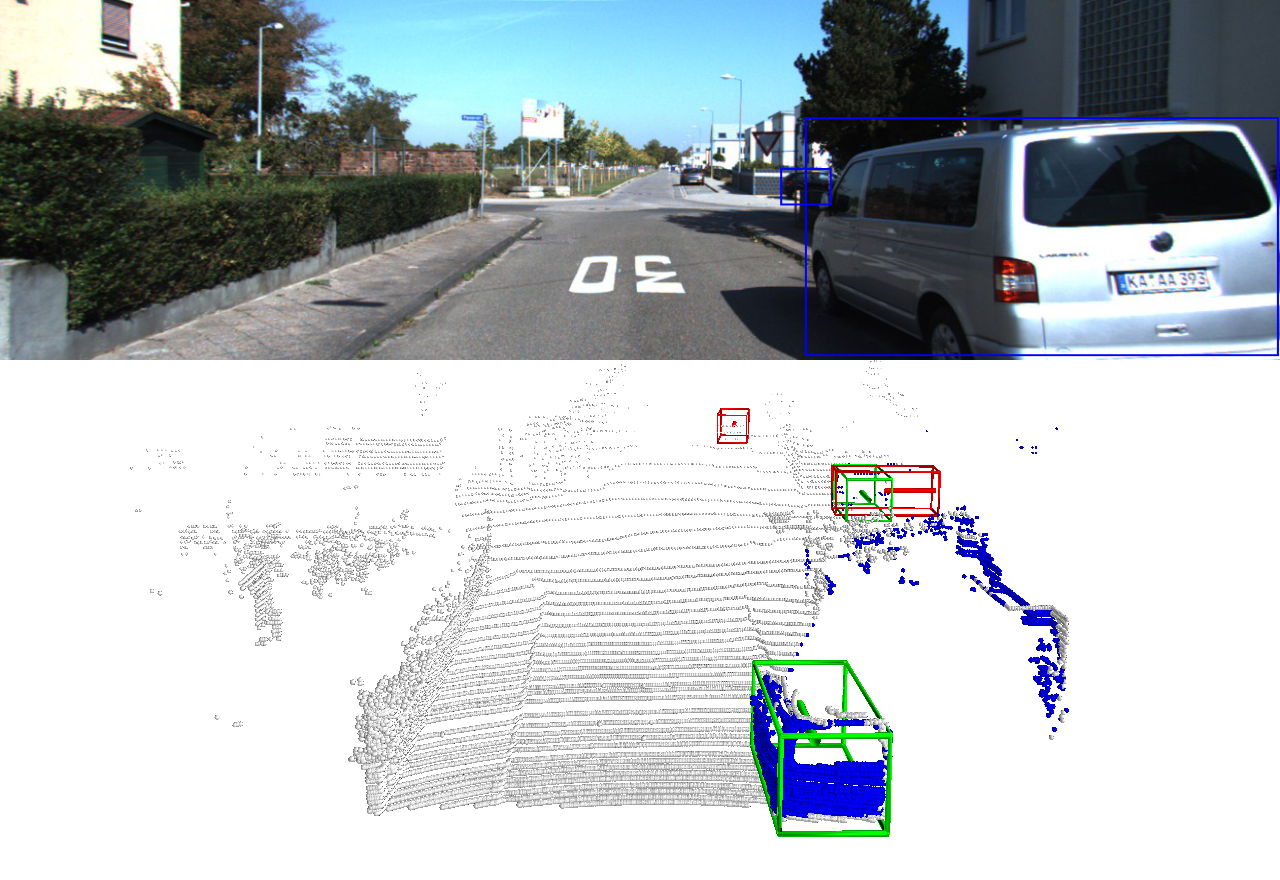}
        &
        \includegraphics[width=0.5\textwidth]{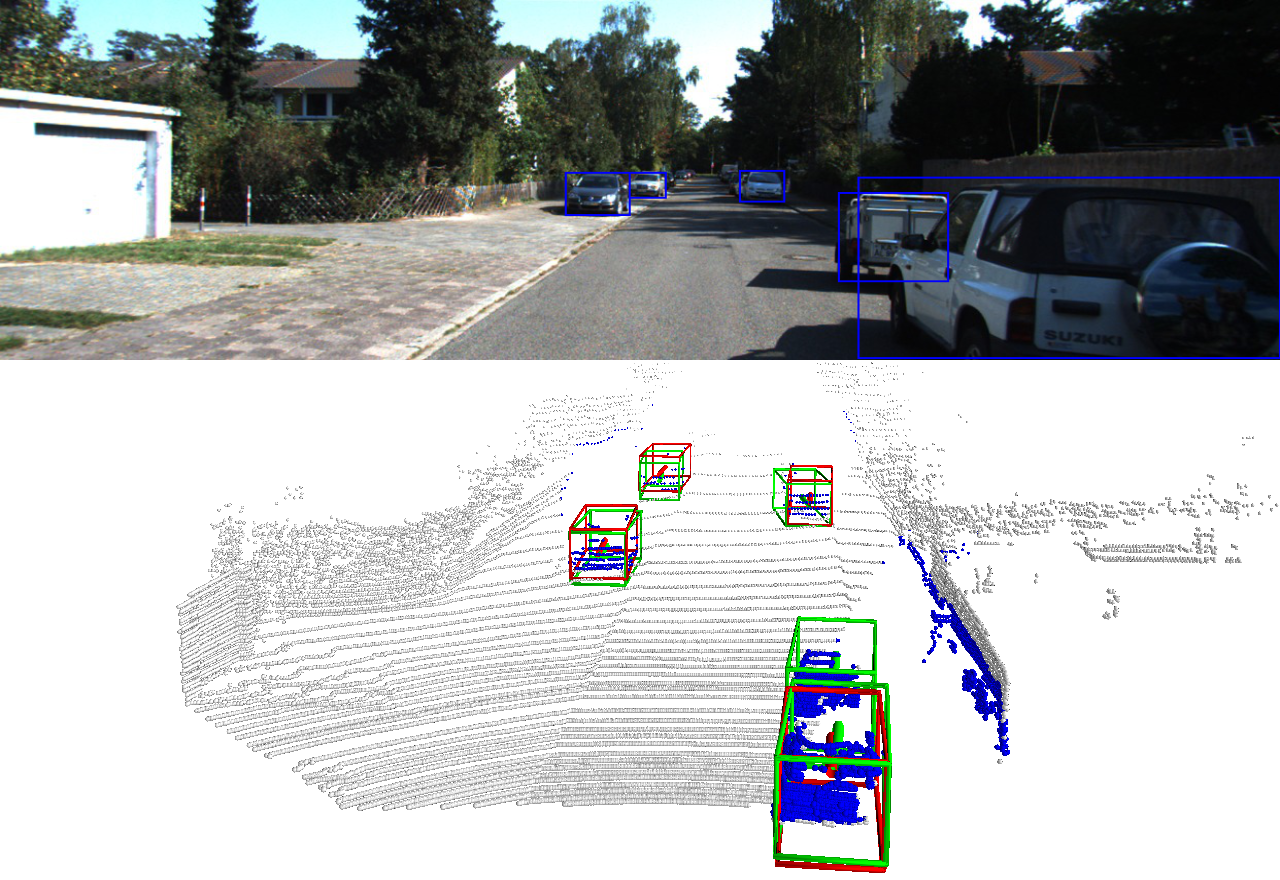}
        \\
        i & j \\
         
        \includegraphics[width=0.5\textwidth]{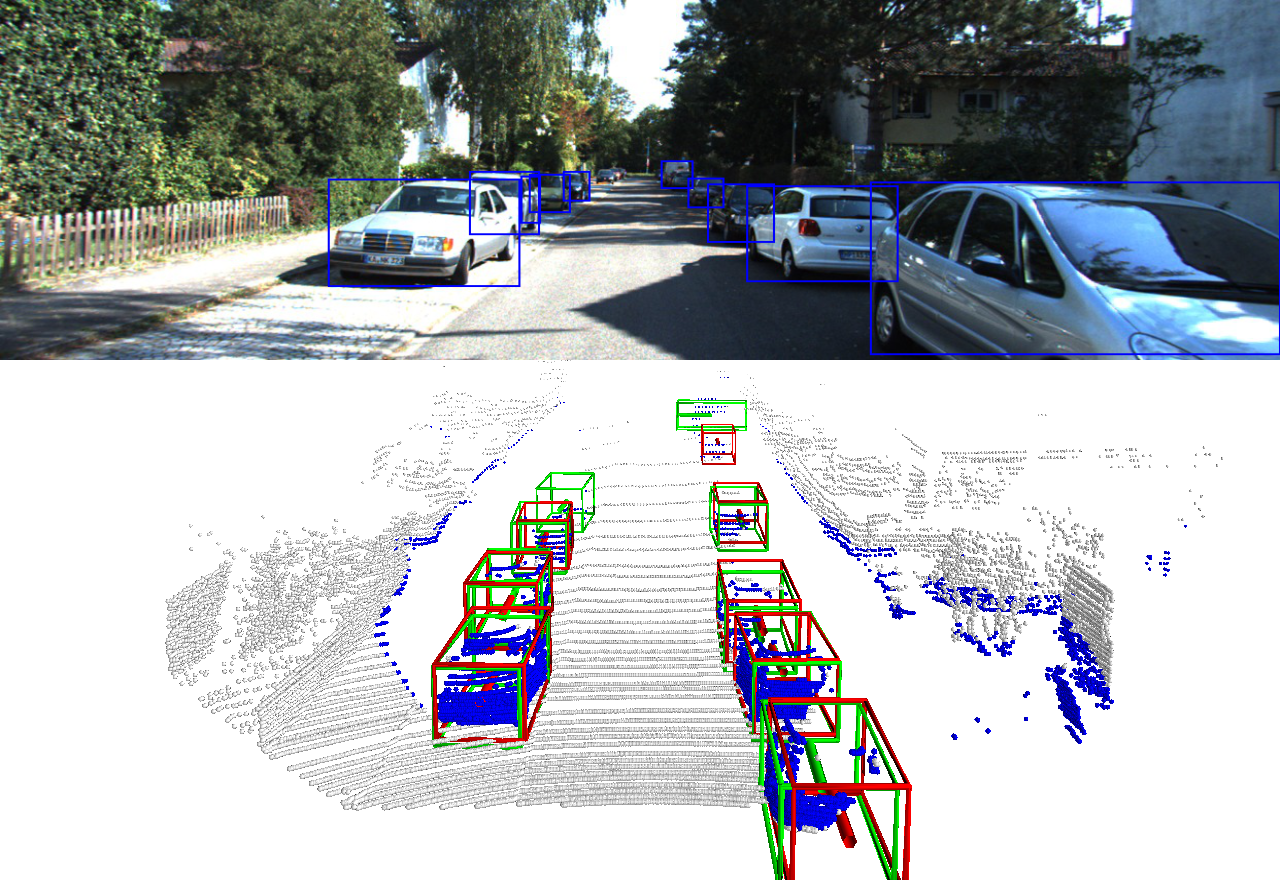}
        &
        \includegraphics[width=0.5\textwidth]{figures/Qualitative_examples/411.png}
        \\
        l & m \\
    \end{tabular}
    \caption{Ours in Green, Ground Truth in Red.}
    \label{f:Qualitative_supp2}
\end{figure}

{\small \bibliographystyle{plain}\bibliography{mccraith_g,vedaldi_general,vedaldi_specific, insafutdinov}}